%% file: main.tex
\documentclass[runningheads]{llncs}
\usepackage[T1]{fontenc}
\usepackage{newtxtext}       %
\usepackage[varvw]{newtxmath}       

\usepackage{multirow}
\usepackage{comment}
\usepackage{booktabs}
\usepackage{tabularx}
\usepackage{tcolorbox}
\tcbuselibrary{breakable}
\usepackage{graphicx}
\usepackage{subcaption}
\usepackage{threeparttable}
\usepackage{balance}

\newtcolorbox{promptbox}{
  breakable,
  colback=gray!5,
  colframe=black!50,
  boxrule=0.5pt,
  arc=2pt,
  left=6pt,
  right=6pt,
  top=6pt,
  bottom=6pt,
  fontupper=\ttfamily\small
}

\begin{document}
\title{LusoClin: A Synthetic Clinical Note Dataset for Medical Information Retrieval in European Portuguese}
\titlerunning{LusoClin: A Synthetic Clinical Note Dataset}

\author{Carla Teixeira Lopes\inst{1,2}\orcidID{0000-0002-4202-791X} \and
Daniel Félix\inst{1,2}\orcidID{0009-0007-1835-2513}}

\authorrunning{C. Teixeira Lopes and D. Félix.}

\institute{Faculty of Engineering of the University of Porto, Porto, Portugal\and INESC TEC, Porto, Portugal
\\
\email{ctl@fe.up.pt,up201905189@up.pt}}

%
%
%
%
\maketitle              
\begin{abstract}
Access to large-scale clinical text collections is crucial for advancing medical information retrieval research, including clinical note search, patient cohort identification, and domain-adapted retrieval models. However, real clinical notes are rarely available due to privacy regulations, which particularly limits progress for low-resource languages such as European Portuguese. 
    We present LusoClin, a publicly released dataset of synthetic clinical notes in European Portuguese, created to support privacy-preserving research on clinical text retrieval. LusoClin was generated by pre-processing real clinical data, fine-tuning the open-source Llama 3.2 (3B) instruction-tuned model using Quantized Low-Rank Adaptation, and applying prompt-based clinical note generation. To facilitate semantic indexing and clinical search, the notes are automatically annotated with clinical entities using a BERT-based model.
    We evaluate LusoClin in terms of utility, fidelity, diversity, and privacy using automatic metrics and human evaluation. Results indicate that the dataset captures clinically relevant language patterns and can support downstream IR applications such as clinical document retrieval and domain-specific model adaptation, although annotators can often distinguish synthetic from real notes and some medical inconsistencies persist. 

\keywords{Clinical Notes \and Synthetic Data Generation \and Large Language Model}
\end{abstract}

\input{sections/introduction}

\input{sections/relatedwork}

\input{sections/methodology}
\input{sections/evaluation}
\input{sections/conclusions}

\begin{credits}
\subsubsection{\ackname} This work was co-financed by Component 5 - Capitalization and Business Innovation, integrated in the Resilience Dimension of the Recovery and Resilience Plan within the scope of the Recovery and Resilience Mechanism (MRR) of the European Union (EU), framed in the Next Generation EU, for the period 2021 - 2026, within project HfPT, with reference 41.
\end{credits}
%
%
%
\bibliographystyle{splncs04}
\bibliography{references}
\end{document}

%% file: sections/introduction.tex
\section{Introduction}

Electronic health records (EHRs) contain longitudinal patient information collected during healthcare delivery~\cite{Cowie2017,Gunter2005}. Much of this information is stored in clinical notes, unstructured records describing clinical encounters~\cite{Oliveira2022}. These notes capture contextual information often absent from structured EHR fields.

Clinical information retrieval plays a central role in enabling access to this unstructured information, useful for decision-making and clinical research~\cite{Sivarajkumar2024}. Retrieval techniques, including recent dense retrieval methods, have also been applied to patient cohort identification by querying electronic health records to identify patient groups with shared criteria~\cite{Jadhav2025,Liu2020,Chamberlin2020}.
Because clinical information is often recorded as unstructured text~\cite{Li2022}, retrieval systems rely on robust text representations, often supported by information extraction and natural language processing techniques~\cite{Shi2023}. In this context, the development and evaluation of clinical IR systems increasingly depend on large-scale, high-quality textual collections, which are difficult to obtain due to privacy regulations~\cite{Long2024}.

This challenge is especially pronounced for languages other than English. While English medical corpora are increasingly available~\cite{Shi2023}, many languages remain severely under-resourced~\cite{Schneider2020}.  Portuguese is a representative example: although some clinical text resources exist for Brazilian Portuguese~\cite{Cabrita2023,Sabia2023,Schneider2021}, European Portuguese remains largely unsupported~\cite{Gloria2024}, primarily due to a lack of publicly available clinical text~\cite{lopes-etal-2019-contributions}. This scarcity limits the development, adaptation, and evaluation of clinical retrieval models for European Portuguese.

Synthetic data is computer-generated data used to augment or replace real data~\cite{ibm_synt_data_definition}. Recent progress in large language models (LLMs) has renewed interest in synthetic data generation showing that LLMs can generate high-quality synthetic clinical text at scale, reducing both cost and privacy concerns~\cite{Long2024,Kweon2023,Waseem2025,Yu2023,Zhang2024}.
In this work, we introduce LusoClin, a large-scale, publicly available dataset of synthetic clinical notes in European Portuguese generated using the open-source Llama 3.2 (3B) instruction-tuned model. Although the dataset presents limitations typical of synthetic clinical text, LusoClin provides a resource for developing and evaluating clinical IR models in European Portuguese, addressing a critical gap in available retrieval resources. The dataset, along with generation prompts, annotations, and human evaluation results, is publicly released in a dedicated repository~\cite{dataset_inesc_repo}.

%% file: sections/relatedwork.tex
\section{Related Work}

Synthetic data has been increasingly explored as a way to address the difficulty of obtaining large-scale real-world data, particularly in sensitive domains such as healthcare. In addition to alleviating data scarcity, synthetic data generation has been shown to offer benefits such as reducing bias~\cite{Kweon2023,Umer2024}, improving performance, scalability, and stronger privacy guarantees~\cite{Joshi2024}. These properties make synthetic data particularly attractive.

Recent advances in LLMs have further accelerated interest in synthetic clinical text generation. Due to extensive pre-training, LLMs exhibit strong language understanding capabilities~\cite{Ding2023,Waseem2025} and can generate fluent unstructured text~\cite{Waseem2025,Xu2024}. As a result, several studies have leveraged LLMs to generate synthetic unstructured clinical datas. While methodologies vary, most approaches share elements related to data preparation, prompt design, and evaluation.

Some studies rely primarily on pre-trained LLMs~\cite{Falis2024,Tang2023}, occasionally augmented with few-shot examples~\cite{Frei2023,Lorge2024,Xu2024}, while others further adapt models through fine-tuning on clinical or task-specific corpora~\cite{Peng2023,Amin-Nejad2020,Kumichev2024}. 

Prompt engineering plays a central role in guiding LLMs to produce clinically meaningful outputs. Prior work has explored diverse prompting strategies, ranging from highly structured prompts specifying content and format constraints~\cite{Falis2024} to iterative prompt refinement~\cite{Tang2023}, few-shot prompting with annotated examples~\cite{Frei2023,Lorge2024}, and frameworks that inject external medical knowledge~\cite{Xu2024}. These studies highlight the importance of balancing guidance and variability to achieve realistic and diverse synthetic clinical text.

Evaluation of synthetic clinical datasets is independent of the generation approach and remains a critical challenge. In the clinical domain, evaluation typically focuses on four key dimensions: utility, fidelity, diversity, and privacy~\cite{Breugel2024}. Utility assesses whether synthetic data can effectively support downstream modelling and retrieval tasks~\cite{Budu2024}, while fidelity measures the degree to which synthetic samples resemble real clinical text in terms of correctness and coherence~\cite{Budu2024,Figueira2022,Long2024}. Diversity reflects the ability to capture clinically meaningful variation~\cite{Eigenschink2023,Long2024}, and privacy ensures that synthetic data cannot be linked back to real individuals~\cite{Eigenschink2023}. These dimensions are often in tension, requiring careful trade-offs during dataset construction and evaluation~\cite{Goncalves2020}.

%% file: sections/methodology.tex
\section{Dataset Construction}
\label{sec:dataset_gen}
We generated LusoClin using Meta’s Llama 3.2~\cite{Dubey2024}, specifically the 3-billion-parameter instruction-tuned variant (``Llama-3.2-3B-Instruct''). This model was selected to support local execution and fine-tuning using publicly available weights, avoiding the need to share sensitive data with third-party services and enabling greater control over the generation. Additional selection criteria included a context window large enough to accommodate the longest clinical documents and example prompts (with 2,048 tokens used as a safe threshold), compatibility with available GPU and storage resources, and support for Portuguese---particularly European Portuguese---within an instruction-based prompt–response framework.

\begin{figure}[h]
\includegraphics[width=1\linewidth]{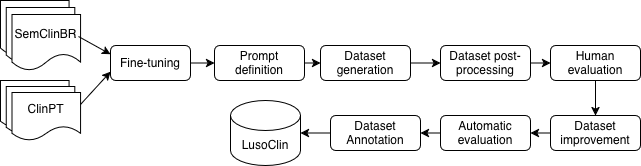}
\caption{Generation pipeline.}
\label{generation_pipeline}
\end{figure}

To ground the generation process in realistic clinical language and structure, we leveraged existing publicly available clinical datasets. Specifically, we used SemClinBr~\cite{Oliveira2022} and CLINpt~\cite{lopes-etal-2019-contributions} as source data. SemClinBr consists of 1,000 Brazilian Portuguese clinical notes spanning four types---admission notes, ambulatory notes, discharge summaries, and nursing notes---and includes annotations for 87 semantic entity types, along with abbreviation and negation labels and two relation types. CLINpt comprises 281 European Portuguese neurology clinical cases with manually annotated named entities, providing complementary coverage of European Portuguese clinical language. Both datasets are anonymized.
An overview of the generation process is shown in Figure~\ref{generation_pipeline}.

\subsection{Fine-tuning}

Following prior work~\cite{Amin-Nejad2020,Peng2023}, we pre-processed both datasets by extracting document text, converting characters to lowercase, normalizing diacritics, and removing punctuation. Fine-tuning was performed using QLoRA~\cite{Dettmers2023} with the AdamW optimizer~\cite{Loshchilov2019}, a learning rate of 3$ e^{-3} $, weight decay of 0.01, four epochs, and a batch size of four. Training used a virtual machine with 100 GB RAM, 100 GB storage, and one NVIDIA RTX 3090 GPU.

\subsection{Prompt definition}

We iteratively refined the generation prompt through empirical experimentation until the model’s outputs consistently resembled realistic clinical notes. Drawing on prior work, we progressively enriched an initial prompt that instructed the model to act as a Portuguese clinician by adding more detailed task descriptions, explicit constraints to mitigate common generation errors, and few-shot examples sampled from SemClinBr. Prompt refinement was accompanied by tuning the model’s temperature.

Prompt constraints controlled aspects such as text length, structure, and style, including avoiding section headers, signatures, lists, repetitive phrasing, and excessive use of generic terms (e.g., repeated mentions of “patient”). These constraints were combined to correct recurring issues while avoiding excessive prompt complexity. We did not explicitly describe the characteristics of each clinical note type, allowing the model to infer them from examples. To encourage lexical and semantic diversity, each prompt instance included an example containing a randomly selected Disease or Syndrome entity from SemClinBr. Attempts to use English-language prompts were unsuccessful.

All prompt templates, associated temperatures, and generated responses from this phase are publicly available in the accompanying repository~\cite{dataset_inesc_repo}. For clarity, we present an English translation of the final prompt template below; the original Portuguese version was used for generation and is included in the repository.

\begin{promptbox}
Write, in European Portuguese, [Clinical Note Type] with 10 sentences, 130 words, and only 1 paragraph. 

Clinical Note Type] must be in running text, without any subdivision. Do not include any ‘Age: 52’-style sections. Don't repeat the word ‘patient’ several times. Don't repeat sentences. Your answer must include only [Clinical Note Type]. 

Some examples are: [Examples with one randomly chosen `Disease or Syndrome' entity]. Don't give any more examples. Don't elaborate on any of these examples.
\end{promptbox}

For each generation prompt, we specified a clinical note type corresponding to those defined in SemClinBr, translated to European Portuguese with the assistance of medical students. Prompts included up to four few-shot examples selected from SemClinBr notes of the requested type that contained a randomly chosen Disease or Syndrome entity; when fewer matching notes were available, all were used. This strategy was adopted to encourage semantic diversity while maintaining note-type consistency.

To mitigate common generation errors, we introduced explicit constraints on length and structure, setting limits of 130 words and 10 sentences based on the average size of SemClinBr notes, and empirically selected a temperature of 0.6. In addition, simple rule-based post-processing was applied to generated outputs to remove instruction-related artifacts, repeated or incomplete sentences, and isolated numbered points.

To assess whether further prompt refinement was necessary before large-scale generation, we conducted a small-scale preliminary evaluation with two final-year medical students. Sixteen synthetic notes (four per note type) and eight real notes from SemClinBr (two per note type) were randomly selected and evenly split between the evaluators. The evaluators were blinded to both the origin and type of each note and completed a Turing-style assessment and qualitative classifications. Although limited in scale, this evaluation was sufficient to determine that no further prompt modifications were required, and the original prompt was retained.

\subsection{Dataset generation and post-processing} \label{ssec:gen_post_proc}

After completing prompt engineering, we generated the full LusoClin dataset comprising 100,000 synthetic clinical notes, with 25,000 notes per clinical note type. Generation was performed on the same virtual machine used for fine-tuning and required a total of 17 days, 20 hours, 50 minutes, and 44 seconds.

Following generation, we applied additional rule-based post-processing to remove noisy outputs. Some generated notes contained instruction-like text, such as requests to generate notes or meta-level guidance, while others resembled transcripts of physician–patient interactions rather than clinical documentation. To filter such cases, we automatically removed notes containing a predefined set of keywords indicative of these errors,  including the portuguese translations of 'to write', 'to rewrite', 'write', 'rewrite','to register', 'you are', formal version of 'you', informal version of 'you'. 

\subsection{Human evaluation}

Human evaluation was conducted due to its importance in assessing the realism (fidelity dimension) and utility of synthetic clinical data. After post-processing, LusoClin comprised 98,571 synthetic clinical notes: 24,759 admission notes, 24,411 ambulatory notes, 24,639 discharge summaries, and 24,762 nursing notes.

The evaluation involved medical professionals and medical students enrolled in the third to sixth years of a six-year medical degree at the University of Porto. Recruitment was conducted via institutional email lists and faculty outreach, and evaluations took place between May 15 and June 2, 2025. Participants completed the evaluation through a dedicated web application\footnote{\url{https://notasclinicas.inesctec.pt}}, where they were shown a clinical note selected with a 90\% probability from LusoClin and a 10\% probability from the real SemClinBr dataset, without being informed of its origin.

For each note, evaluators indicated whether they believed it to be real or synthetic and optionally provided justifications and excerpt-level comments. To increase reliability while preserving coverage, we collected up to two independent evaluations per note using a controlled random assignment strategy that avoided repeated evaluations by the same user.
The results of this evaluation are included in Section~\ref{humanEvaluation}.

\subsection{Dataset improvements}

Following the initial evaluation, we applied a series of refinements to improve LusoClin, focusing on reducing Brazilian Portuguese (PT-BR) influence, correcting syntactic issues, and removing irrelevant content. To identify PT-BR usage, we employed PtVId~\cite{Sousa2025}, a BERT-based classifier that distinguishes between European Portuguese (PT-PT) and PT-BR and provides a confidence score. In an initial pass, only 31,525 of the 98,564 notes (approximately 32\%) were classified as PT-PT. To mitigate this, we manually analyzed the 1,000 most salient terms (by TF–IDF) in notes classified as PT-BR and replaced PT-BR expressions with their PT-PT equivalents. This process resulted in modifications to 71,381 notes and increased the number of notes classified as PT-PT to 56,079 (approximately 57\%). 

We then addressed recurrent syntactic issues, primarily the absence of diacritics, which led to corrections in 60,590 clinical notes. These corrections improved linguistic quality without altering clinical content.
Finally, we removed notes deemed irrelevant. This included six notes containing no Portuguese content and 2,362 notes containing terms indicative of prompt echoing or generic descriptions rather than clinical documentation. After these filtering steps, LusoClin comprised 96,202 clinical notes, of which 54,518 (approximately 57\%) were classified as PT-PT.

Given our objective of producing a strictly European Portuguese clinical dataset, we retained only notes classified as PT-PT. The final version of LusoClin contains 54,518 synthetic clinical notes: 14,253 admission notes, 13,561 ambulatory notes, 9,032 discharge summaries, and 17,672 nursing notes. 

\subsection{Automatic Evaluation}
\label{sec:met_automaticEvaluation}

To complement human evaluation, we conducted an automatic assessment targeting utility, fidelity, diversity, and privacy. Evaluation included perplexity, textual and entity-level statistics, t-SNE visualizations, Central Moment Discrepancy (CMD), cosine similarity, and Hausdorff distance to compare synthetic and real clinical text representations.

\subsection{Dataset Annotation}

To enable entity-level evaluation, we automatically annotated LusoClin with named entities using MediAlbertina PT-PT~\cite{Nunes2024}, a 1.5B BERT-based model fine-tuned for NER. The model was pre-trained on European Portuguese text~\cite{Rodrigues2023} and further adapted using medical documents from Portugal's largest public hospital~\cite{Nunes2024}. We applied the publicly available NER version to all clinical notes using the entity categories Diagnosis, Symptom, Medication/Drug, Dosage, Medical Procedure, Vital Sign, Result, and Progress.

%% file: sections/evaluation.tex
\section{Evaluation Results}
This section reports the results of our two-stage evaluation.

\subsection{Human evaluation results}
\label{humanEvaluation}

We collected 149 evaluations of 78 clinical notes. Among these, 71 notes (57 LusoClin and 14 SemClinBr) received two evaluations, summarized in Table~\ref{two_eval_dist}.

\begin{table}
\centering
\caption{Distribution of votes for clinical notes with two independent evaluations (R = Real, S = Synthetic).}
\label{two_eval_dist}

\begin{tabularx}{\linewidth}{@{}l *{6}{>{\centering\arraybackslash}X}@{}}
\toprule
\multirow{2}{*}{\textbf{Note type}} &
\multicolumn{3}{c}{\textbf{LusoClin}} &
\multicolumn{3}{c}{\textbf{SemClinBr}} \\
\cmidrule(lr){2-4} \cmidrule(lr){5-7}
& \textbf{2R} & \textbf{1R+1S} & \textbf{2S} & \textbf{2R} & \textbf{1R+1S} & \textbf{2S} \\
\midrule
Admission notes     & 0 & 2  & 6  & 0 & 1 & 0 \\
Ambulatory notes    & 1 & 10 & 9  & 1 & 4 & 2 \\
Discharge summaries & 1 & 8  & 8  & 0 & 2 & 0 \\
Nursing notes       & 1 & 2  & 9  & 1 & 1 & 2 \\
\midrule
\textbf{Total}      & \textbf{3} & \textbf{22} & \textbf{32} & \textbf{2} & \textbf{8} & \textbf{4} \\
\bottomrule
\end{tabularx}
\end{table}

We computed Cohen’s Kappa~\cite{Cohen1960}, obtaining a value of -0.04, indicating no meaningful inter-annotator agreement~\cite{McHugh2012,Sim2005,Viera2005}. This may reflect the difficulty of distinguishing synthetic from real clinical notes or differences in evaluator criteria~\cite{Sim2005}. We therefore restrict the remaining analysis to the 41 notes with concordant evaluations.

Table~\ref{turing_test_results} shows the human classifications. Only 9\% of LusoClin notes were classified as real, while 67\% of SemClinBr notes were judged synthetic. This suggests evaluators adopted strict realism criteria, potentially interpreting characteristics of genuine clinical notes as indicators of synthetic origin. 
Five LusoClin notes received concordant justifications, attributed to medical errors, meta-data artifacts, missing key information, or an overall synthetic appearance. Discordant justifications are summarized in Table~\ref{synthetic_votes_reasons}. Although most justifications were provided by only one evaluator, they remain useful for identifying potential weaknesses and informing dataset improvements.

\begin{table}
\centering
\caption{Human classification of clinical notes.}
\label{turing_test_results}

\begin{tabularx}{\linewidth}{@{}l *{4}{>{\centering\arraybackslash}X}@{}}
\toprule
\multirow{2}{*}{\textbf{Note type}} &
\multicolumn{2}{c}{\textbf{LusoClin}} &
\multicolumn{2}{c}{\textbf{SemClinBr}} \\
\cmidrule(lr){2-3} \cmidrule(lr){4-5}
& \textbf{Real} & \textbf{Synthetic} & \textbf{Real} & \textbf{Synthetic} \\
\midrule
Admission notes     & 0 (0\%)  & 6 (100\%) & 0 (0\%)  & 0 (0\%)  \\
Ambulatory notes    & 1 (10\%) & 9 (90\%)  & 1 (33\%) & 2 (67\%) \\
Discharge summaries & 1 (13\%) & 8 (87\%)  & 0 (0\%)  & 0 (0\%)  \\
Nursing notes       & 1 (10\%) & 9 (90\%)  & 1 (33\%) & 2 (67\%) \\
\midrule
\textbf{Total}      & \textbf{3 (9\%)} & \textbf{32(91\%)} & \textbf{2 (33\%)} & \textbf{4 (67\%)} \\
\bottomrule
\end{tabularx}
\end{table}


\begin{table}[t]
\centering
\caption{Number and proportion of justifications for classifying notes as synthetic, excluding empty justifications.}
\label{synthetic_votes_reasons}

\begin{tabular}{@{}l @{\hspace{0.5cm}} c @{\hspace{0.4cm}} c @{}p{0.7cm}@{} c @{\hspace{0.4cm}} c@{}}
\toprule
\multirow{2}{*}{\textbf{Justification}} &
\multicolumn{2}{c}{\textbf{LusoClin}} &
&
\multicolumn{2}{c}{\textbf{SemClinBr}} \\
\cmidrule(lr){2-3} \cmidrule(lr){5-6}
& \textbf{N} & \textbf{\%} & & \textbf{N} & \textbf{\%} \\
\midrule
Feels synthetic                    & 7 & 17.07\% & & 0 & 0.00\% \\
Medical errors                     & 7 & 17.07\% & & 1 & 50.00\% \\
Missing key information            & 6 & 14.63\% & & 0 & 0.00\% \\
Non-sensical                       & 5 & 12.20\% & & 0 & 0.00\% \\
Contains meta-data                 & 4 & 9.76\%  & & 0 & 0.00\% \\
Repetitive                         & 3 & 7.32\%  & & 0 & 0.00\% \\
Incorrect structure and/or grammar & 3 & 7.32\%  & & 0 & 0.00\% \\
Unusual terminology                & 3 & 7.32\%  & & 0 & 0.00\% \\
Brazilian words                    & 2 & 4.88\%  & & 1 & 50.00\% \\
Non-existent medical terminology   & 1 & 2.44\%  & & 0 & 0.00\% \\
\bottomrule
\end{tabular}
\end{table}

\subsection{Automatic evaluation results}

As described in Section~\ref{sec:met_automaticEvaluation}, we combined several automatic evaluation methods targeting utility, fidelity, diversity, and privacy. Table~\ref{tab:eval_summary} summarizes the methods and baselines used.

\begin{table}[t]
\centering
\caption{Evaluation methods, assessed dimensions, and comparison baselines.}
\label{tab:eval_summary}

\begin{threeparttable}
\begin{tabular}{@{}l @{\hspace{0.7cm}} l @{\hspace{0.7cm}} l@{}}
\toprule
\textbf{Method} & \textbf{Dimensions} & \textbf{Baseline} \\
\midrule
Perplexity & Utility, fidelity & SemClinBr \\
Statistics & Utility, fidelity & SemClinBr \\
t-SNE plot\tnote{*} & Utility, fidelity, diversity & \cite{Xu2024} \\
CMD\tnote{*} & Fidelity & \cite{Xu2024} \\
Cosine similarity\tnote{*} & Diversity, privacy & SemClinBr, \cite{Hernadez2023} \\
Hausdorff distance\tnote{*} & Privacy & \cite{Hernadez2023} \\
\bottomrule
\end{tabular}

\begin{tablenotes}
\footnotesize
\item[*] Computed on a subset of LusoClin matched to the size of SemClinBr.
\end{tablenotes}
\end{threeparttable}
\end{table}

Perplexity and textual statistics were computed on the full dataset, while t-SNE, CMD, cosine similarity, and Hausdorff distance used a subset of 1,000 LusoClin notes matched to SemClinBr. These methods relied on multilingual SentenceBERT embeddings~\cite{SentenceBERT}. SemClinBr served as baseline for perplexity, statistics, and cosine similarity, while~\cite{Xu2024} provided the English-language baseline for t-SNE and CMD.

\subsubsection{Perplexity}

We computed the average perplexity of GlórIA~\cite{Gloria2024} and Cabrita~\cite{Cabrita2023} on LusoClin and SemClinBr. LusoClin achieved lower perplexity on both models (68.32 vs.\ 82.57 for GlórIA; 29.44 vs.\ 37.31 for Cabrita), indicating that the synthetic notes were easier to predict. Since lower perplexity has been associated with medically coherent clinical content~\cite{Amin-Nejad2020,Kweon2023,Melamud2019}, these results support the utility and fidelity of LusoClin.

\subsubsection{Textual and entity statistics}

\begin{figure}[t]
\centering
\includegraphics[width=0.7\columnwidth]{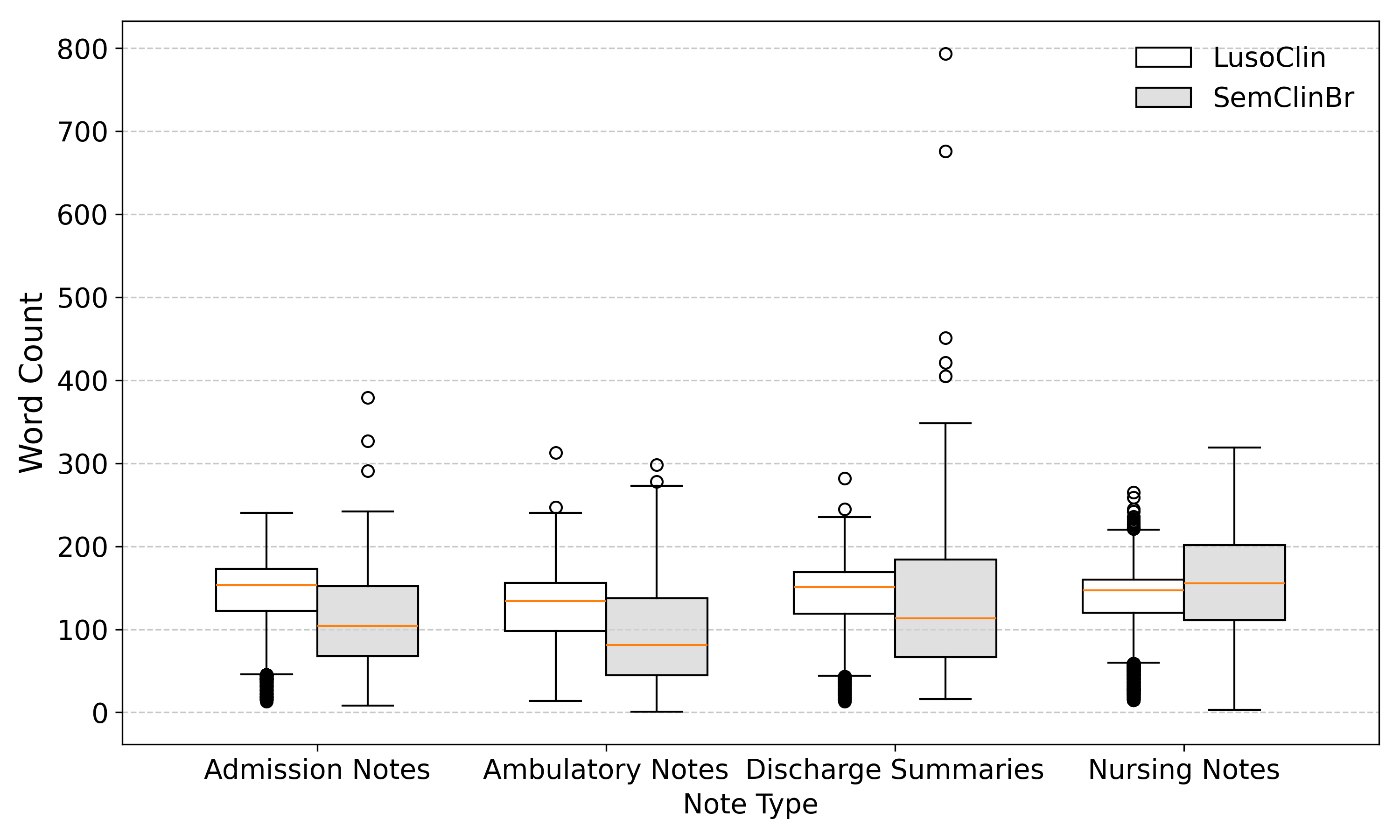}
\caption{Word count distributions for each note type.}
\label{fig:word_count_comparison}
\end{figure}

\begin{figure}[t]
\centering
\includegraphics[width=0.7\columnwidth]{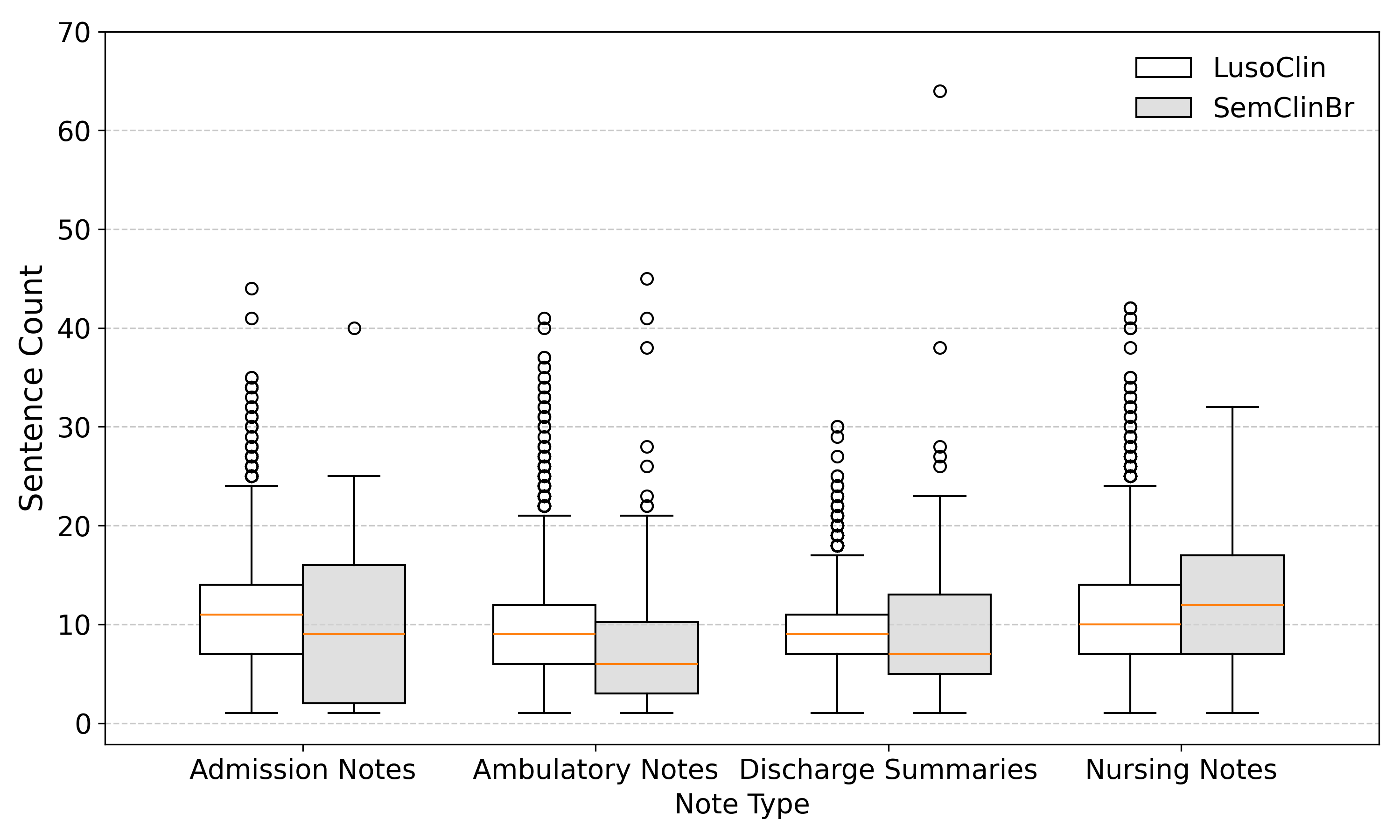}
\caption{Sentence count distributions for each note type.}
\label{fig:sentence_count_comparison}
\end{figure}

\begin{figure}[t]
\centering
\includegraphics[width=0.7\columnwidth]{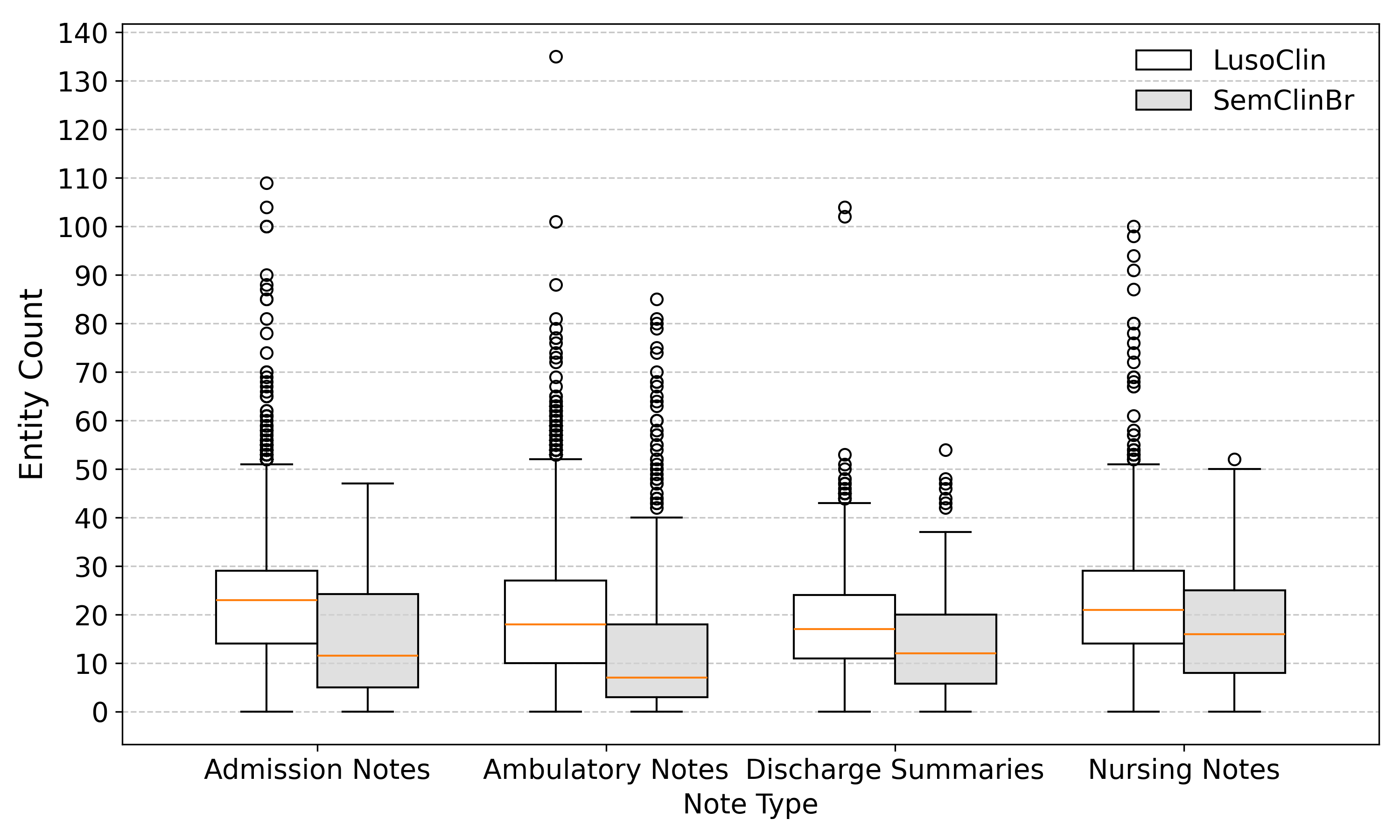}
\caption{Entity count distributions for each note type.}
\label{fig:entity_count_comparison}
\end{figure}

We compared textual and entity-level statistics between datasets, including document length, sentence count, and entity distributions. Figure~\ref{fig:word_count_comparison} shows that synthetic notes are slightly longer than real notes, while Figure~\ref{fig:sentence_count_comparison} indicates similar sentence structures, although SemClinBr displays greater variability. Table~\ref{entity_type_count_percentage} reports entity type frequencies. The average absolute difference between datasets was 3.71\%, indicating close alignment in medical entity distributions and supporting the realism of LusoClin. Figure~\ref{fig:entity_count_comparison} further shows that synthetic notes contain more entities on average, suggesting denser medical terminology.

\begin{table}
\centering
\caption{Entity type frequencies.}
\label{entity_type_count_percentage}

\begin{tabular}{@{}l @{\hspace{0.5cm}} c @{\hspace{0.4cm}} c @{}p{0.7cm}@{} c @{\hspace{0.4cm}} c@{}}
\toprule
\multirow{2}{*}{\textbf{Entity type}} &
\multicolumn{2}{c}{\textbf{LusoClin}} &
&
\multicolumn{2}{c}{\textbf{SemClinBr}} \\
\cmidrule(lr){2-3} \cmidrule(lr){5-6}
& \textbf{Count} & \textbf{\%} & & \textbf{Count} & \textbf{\%} \\
\midrule
Diagnosis           & 273,520 & 24.28\% & & 2,378 & 16.31\% \\
Symptom             & 193,152 & 17.14\% & & 1,915 & 13.14\% \\
Medical Procedure   & 189,966 & 16.86\% & & 2,228 & 15.28\% \\
Result              & 153,159 & 13.59\% & & 2,616 & 17.94\% \\
Vital Sign          & 142,933 & 12.69\% & & 2,617 & 17.95\% \\
Medication/Drug     & 85,835  & 7.62\%  & & 1,331 & 9.13\%  \\
Dosage              & 57,143  & 5.07\%  & & 1,281 & 8.79\%  \\
Progress            & 30,939  & 2.75\%  & & 212   & 1.45\%  \\
\bottomrule
\end{tabular}
\end{table}

\subsubsection{t-SNE plots}

\begin{figure}
\centering
\includegraphics[scale=0.5]{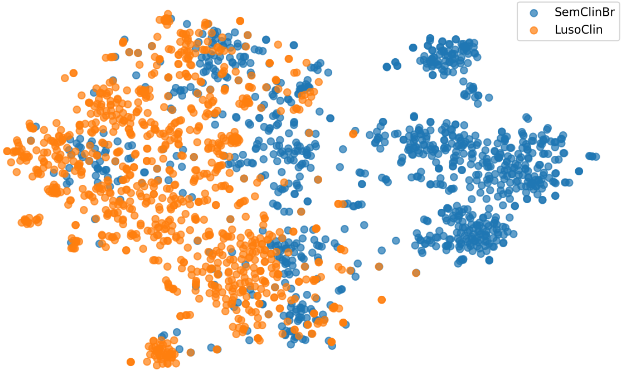}
\caption{t-SNE plots of LusoClin and SemClinBr.}
\label{t_sne_plot_results}
\end{figure}

We examined t-SNE visualizations of the LusoClin subset and SemClinBr, shown in Figure~\ref{t_sne_plot_results}. The plots reveal only partial overlap between the two datasets, suggesting that the distribution of LusoClin could more closely align with that of SemClinBr to achieve comparable levels of utility and fidelity. This observation is consistent with findings reported by Xu et al.~\cite{Xu2024}, whose English-language baseline exhibits a greater overlap between synthetic and real data distributions.
At the same time, the overall spacing and dispersion of LusoClin data points appear similar to those of SemClinBr, mirroring the behavior observed in the English baseline~\cite{Xu2024}. This indicates that LusoClin captures a comparable level of diversity, which is a positive outcome given that SemClinBr is a multi-specialty clinical dataset.

\subsubsection{Central Moment Discrepancy}

We additionally computed the average CMD between the LusoClin subset and SemClinBr, obtaining a value of 0.395. Compared with the English-language baseline reported by Xu et al.~\cite{Xu2024}, LusoClin achieves a lower CMD on this metric. This result indicates limited shift between the synthetic and real clinical notes, providing further evidence that LusoClin preserves key characteristics of the clinical domain and exhibits satisfactory fidelity.

\subsubsection{Cosine similarity}
We computed cosine similarity using sentence-level embeddings to assess both diversity and privacy. First, we calculated the average cosine similarity over all pairwise combinations of sentences within the LusoClin subset. LusoClin exhibits a lower average pairwise cosine similarity (0.091) than SemClinBr (0.137), indicating higher internal diversity among synthetic notes. This is a positive outcome, as it suggests that LusoClin is not overly repetitive. One possible concern is that increased diversity could reflect a shift away from the clinical domain toward more general language; however, this interpretation is not supported by the human evaluation and CMD results, which indicate preservation of clinical characteristics.

For privacy, we then computed the average cosine similarity between cross-dataset pairs consisting of one LusoClin sample and one SemClinBr sample, considering (1) all note types jointly and (2) pairs matched by note type. As shown in Table~\ref{cosine_similarity_hausD_results}, cross-dataset similarities are consistently low. Following the privacy preservation criterion proposed by Henrnadez et al.~\cite{Hernadez2023}, which sets a cosine similarity threshold of 0.5, both the type-agnostic and type-matched results fall well below this threshold. This provides strong evidence that LusoClin does not exhibit excessive similarity to real clinical notes and therefore supports its privacy-preserving nature.

\begin{table}
\centering
\caption{Average cosine similarity (CosineS) and Hausdorff distance (HausD) between pairs consisting of one LusoClin note and one SemClinBr note.}
\label{cosine_similarity_hausD_results}

\begin{tabularx}{\linewidth}{@{}l >{\centering\arraybackslash}X >{\centering\arraybackslash}X@{}}
\toprule
\textbf{Note type} & \textbf{CosineS} & \textbf{HausD} \\
\midrule
All types           & 0.052 & 1.272 \\
Admission notes     & 0.061 & 1.267 \\
Ambulatory notes    & 0.049 & 1.274 \\
Discharge summaries & 0.068 & 1.275 \\
Nursing notes       & 0.073 & 1.244 \\
\bottomrule
\end{tabularx}
\end{table}

\subsubsection{Hausdorff distance}

We computed the average Hausdorff distance between cross-dataset pairs consisting of one LusoClin note and one SemClinBr note, considering both all note types jointly and pairs matched by note type. These results were compared against the privacy preservation baseline proposed by Hernadez et al.~\cite{Hernadez2023}, which sets a threshold value of 1. As shown in Table~\ref{cosine_similarity_hausD_results}, all computed Hausdorff distances exceed this threshold, both in the type-agnostic and type-matched settings. This provides additional evidence that synthetic notes are sufficiently separated from real notes, supporting the privacy-preserving properties of LusoClin.

%% file: sections/conclusions.tex
\section{Conclusions}

Based on the applied evaluation methods, we draw several conclusions regarding the quality of LusoClin.
Utility and fidelity were the most critical dimensions and were primarily assessed through human evaluation. While evaluators identified issues such as nonsensical content and medical errors in a subset of notes, many synthetic notes were medically coherent and informative. Although most LusoClin notes did not pass as real, this outcome must be interpreted in light of the fact that two-thirds of the SemClinBr notes were also judged synthetic. This highlights the inherent difficulty and subjectivity of human evaluation in synthetic data settings, where evaluators tend to apply stricter realism criteria.

Automatic evaluations of utility and fidelity yielded encouraging results, with LusoClin matching baselines on t-SNE plot and textual and entity statistics and outperforming them on perplexity and CMD. Overall, these findings indicate that LusoClin provides a reasonable level of utility and fidelity, though quality issues remain in a subset of notes.

Evaluation of diversity showed that LusoClin is at least as diverse as SemClinBr, suggesting that it meets diversity requirements for downstream use. Privacy assessment results were also positive. 
The combination of our evaluation results and the use of professionally de-identified source data strongly suggests that LusoClin does not pose privacy risks.

Overall, the results indicate that LusoClin cannot fully replace real clinical data but can serve as a valuable complementary resource for research, particularly in low-resource settings.

%% file: references.bib
@article{Sivarajkumar2024,
   author = {Sonish Sivarajkumar and Haneef Ahamed Mohammad and David Oniani and Kirk Roberts and William Hersh and Hongfang Liu and Daqing He and Shyam Visweswaran and Yanshan Wang},
   doi = {10.1007/S41666-024-00159-4},
   issn = {2509498X},
   issue = {2},
   journal = {Journal of Healthcare Informatics Research},
   month = {6},
   pages = {313-352},
   publisher = {Springer Science and Business Media Deutschland GmbH},
   title = {Clinical Information Retrieval: A Literature Review},
   volume = {8},
   url = {https://link.springer.com/article/10.1007/s41666-024-00159-4},
   year = {2024}
}

@article{Chamberlin2020,
   author = {Steven R. Chamberlin and Steven D. Bedrick and Aaron M. Cohen and Yanshan Wang and Andrew Wen and Sijia Liu and Hongfang Liu and William R. Hersh},
   doi = {10.1093/JAMIAOPEN/OOAA026},
   issn = {25742531},
   issue = {3},
   journal = {JAMIA Open},
   month = {10},
   pages = {395-404},
   publisher = {Oxford Academic},
   title = {Evaluation of patient-level retrieval from electronic health record data for a cohort discovery task},
   volume = {3},
   url = {https://dx.doi.org/10.1093/jamiaopen/ooaa026},
   year = {2020}
}

@misc{Jadhav2025,
      title={Cohort Retrieval using Dense Passage Retrieval}, 
      author={Pranav Jadhav},
      year={2025},
      eprint={2507.01049},
      archivePrefix={arXiv},
      primaryClass={cs.IR},
      url={https://arxiv.org/abs/2507.01049}, 
}

@article{Liu2020,
   author = {Sijia Liu and Yanshan Wang and Andrew Wen and Liwei Wang and Na Hong and Feichen Shen and Steven Bedrick and William Hersh and Hongfang Liu},
   doi = {10.2196/17376},
   issn = {2291-9694},
   issue = {10},
   journal = {JMIR Medical Informatics},
   month = {10},
   publisher = {JMIR Publications Inc.},
   title = {Implementation of a Cohort Retrieval System for Clinical Data Repositories Using the Observational Medical Outcomes Partnership Common Data Model: Proof-of-Concept System Validation},
   volume = {8},
   year = {2020}
}

@article{Shi2023,
    author={Shi, Jiyun
    and Yuan, Zhimeng
    and Guo, Wenxuan
    and Ma, Chen
    and Chen, Jiehao
    and Zhang, Meihui},
    title={Knowledge-graph-enabled biomedical entity linking: a survey},
    journal={World Wide Web},
    year={2023},
    month={Sep},
    day={01},
    volume={26},
    number={5},
    pages={2593-2622},
    issn={1573-1413},
    note={doi: 10.1007/s11280-023-01144-4},
    url = {https://doi.org/10.1007/s11280-023-01144-4}
}

@misc{Schneider2020,
    title={BioBERTpt - A Portuguese Neural Language Model for Clinical Named Entity Recognition},
    author={Elisa Terumi Rubel Schneider and João Vitor Andrioli de Souza and Julien Knafou and Lucas Emanuel Silva e Oliveira and Jenny Copara and Yohan Bonescki Gumiel and Lucas Ferro Antunes de Oliveira and Emerson Cabrera Paraiso and Douglas Teodoro and Cláudia Maria Cabral Moro},
    year={2020},
    booktitle={Proceedings of the 3rd Clinical Natural Language Processing Workshop},
    pages={65--72},
    publisher={Association for Computational Linguistics},
    doi={10.18653/v1/2020.clinicalnlp-1.7},
    url={https://aclanthology.org/2020.clinicalnlp-1.7/},
}

@article{Oliveira2022,
    author={Oliveira, Lucas Emanuel Silva e.
    and Peters, Ana Carolina
    and da Silva, Adalniza Moura Pucca
    and Gebeluca, Caroline Pilatti
    and Gumiel, Yohan Bonescki
    and Cintho, Lilian Mie Mukai
    and Carvalho, Deborah Ribeiro
    and Al Hasan, Sadid
    and Moro, Claudia Maria Cabral},
    title={SemClinBr - a multi-institutional and multi-specialty semantically annotated corpus for Portuguese clinical NLP tasks},
    journal={Journal of Biomedical Semantics},
    year={2022},
    month={May},
    day={08},
    volume={13},
    number={1},
    pages={13},
    issn={2041-1480},
    note={doi: 10.1186/s13326-022-00269-1},
    url={https://doi.org/10.1186/s13326-022-00269-1}
}

@inproceedings{lopes-etal-2019-contributions,
    title = "Contributions to Clinical Named Entity Recognition in {P}ortuguese",
    author = "Lopes, F{\'a}bio  and
      Teixeira, C{\'e}sar  and
      Gon{\c{c}}alo Oliveira, Hugo",
    editor = "Demner-Fushman, Dina  and
      Cohen, Kevin Bretonnel  and
      Ananiadou, Sophia  and
      Tsujii, Junichi",
    booktitle = "Proceedings of the 18th BioNLP Workshop and Shared Task",
    month = aug,
    year = "2019",
    address = "Florence, Italy",
    publisher = "Association for Computational Linguistics",
    url = "https://aclanthology.org/W19-5024",
    note = "doi: 10.18653/v1/W19-5024",
    pages = "223--233",
}

@InProceedings{Rodrigues2023,
    author="Rodrigues, Jo{\~a}o
    and Gomes, Lu{\'i}s
    and Silva, Jo{\~a}o
    and Branco, Ant{\'o}nio
    and Santos, Rodrigo
    and Cardoso, Henrique Lopes
    and Os{\'o}rio, Tom{\'a}s",
    editor="Moniz, Nuno
    and Vale, Zita
    and Cascalho, Jos{\'e}
    and Silva, Catarina
    and Sebasti{\~a}o, Raquel",
    title="Advancing Neural Encoding of Portuguese with Transformer Albertina PT-*",
    booktitle="Progress in Artificial Intelligence",
    year="2023",
    publisher="Springer Nature Switzerland",
    address="Cham",
    pages="441--453",
    isbn="978-3-031-49008-8",
    note="doi: 10.1007/978-3-031-49008-8\_35"
}

@misc{ibm_synt_data_definition,
  author = {{IBM}},
  title = {What is synthetic data?},
  howpublished = {\url{https://research.ibm.com/blog/what-is-synthetic-data}},
  note = {Accessed: 2024-08-27}
}

@article{Umer2024,
    author={Umer, Fahad
    and Adnan, Niha},
    title={Generative artificial intelligence: synthetic datasets in dentistry},
    journal={BDJ Open},
    year={2024},
    month={Mar},
    day={01},
    volume={10},
    number={1},
    pages={13},
    issn={2056-807X},
    note={doi: 10.1038/s41405-024-00198-4},
    url={https://doi.org/10.1038/s41405-024-00198-4}
}

@misc{Kweon2023,
      title={Publicly Shareable Clinical Large Language Model Built on Synthetic Clinical Notes}, 
      author={Sunjun Kweon and Junu Kim and Jiyoun Kim and Sujeong Im and Eunbyeol Cho and Seongsu Bae and Jungwoo Oh and Gyubok Lee and Jong Hak Moon and Seng Chan You and Seungjin Baek and Chang Hoon Han and Yoon Bin Jung and Yohan Jo and Edward Choi},
      year={2024},
      eprint={2309.00237},
      archivePrefix={arXiv},
      primaryClass={cs.CL},
      note={doi: 10.48550/arXiv.2309.00237}
}

@article{Joshi2024,
  author={Joshi, Indu and Grimmer, Marcel and Rathgeb, Christian and Busch, Christoph and Bremond, Francois and Dantcheva, Antitza},
  journal={IEEE Transactions on Pattern Analysis and Machine Intelligence}, 
  title={Synthetic Data in Human Analysis: A Survey}, 
  year={2024},
  volume={46},
  number={7},
  pages={4957-4976},
  note={doi: 10.1109/TPAMI.2024.3362821}
}

@misc{Long2024,
    title={On LLMs-Driven Synthetic Data Generation, Curation, and Evaluation: A Survey}, 
    author={Lin Long and Rui Wang and Ruixuan Xiao and Junbo Zhao and Xiao Ding and Gang Chen and Haobo Wang},
    year={2024},
    eprint={2406.15126},
    archivePrefix={arXiv},
    primaryClass={cs.CL},
    url={https://arxiv.org/abs/2406.15126}, 
    note={doi: 10.48550/arXiv.2406.15126},
}

@misc{Ding2023,
    title={Is GPT-3 a Good Data Annotator?}, 
    author={Bosheng Ding and Chengwei Qin and Linlin Liu and Yew Ken Chia and Shafiq Joty and Boyang Li and Lidong Bing},
    year={2023},
    eprint={2212.10450},
    archivePrefix={arXiv},
    primaryClass={cs.CL},
    url={https://arxiv.org/abs/2212.10450},
    note={doi: 10.48550/arXiv.2212.10450}
}

@inproceedings{Yu2023,
    author = {Yu, Yue and Zhuang, Yuchen and Zhang, Jieyu and Meng, Yu and Ratner, Alexander and Krishna, Ranjay and Shen, Jiaming and Zhang, Chao},
    title = {Large language model as attributed training data generator: a tale of diversity and bias},
    year = {2023},
    publisher = {Curran Associates Inc.},
    address = {Red Hook, NY, USA},
    booktitle = {Proceedings of the 37th International Conference on Neural Information Processing Systems},
    articleno = {2433},
    numpages = {51},
    location = {New Orleans, LA, USA},
    series = {NIPS '23}
    }

@Article{Figueira2022,
    AUTHOR = {Figueira, Alvaro and Vaz, Bruno},
    TITLE = {Survey on Synthetic Data Generation, Evaluation Methods and GANs},
    JOURNAL = {Mathematics},
    VOLUME = {10},
    YEAR = {2022},
    NUMBER = {15},
    ARTICLE-NUMBER = {2733},
    URL = {https://www.mdpi.com/2227-7390/10/15/2733},
    ISSN = {2227-7390},
    NOTE = {doi: 10.3390/math10152733}
}

@article{Goncalves2020,
    author={Goncalves, Andre
    and Ray, Priyadip
    and Soper, Braden
    and Stevens, Jennifer
    and Coyle, Linda
    and Sales, Ana Paula},
    title={Generation and evaluation of synthetic patient data},
    journal={BMC Medical Research Methodology},
    year={2020},
    month={May},
    day={07},
    volume={20},
    number={1},
    pages={108},
    issn={1471-2288},
    note={doi: 10.1186/s12874-020-00977-1},
    url={https://doi.org/10.1186/s12874-020-00977-1}
}

@article{Zhang2024,
author={Zhang, Gongbo
and Jin, Qiao
and Zhou, Yiliang
and Wang, Song
and Idnay, Betina
and Luo, Yiming
and Park, Elizabeth
and Nestor, Jordan G.
and Spotnitz, Matthew E.
and Soroush, Ali
and Campion, Thomas R.
and Lu, Zhiyong
and Weng, Chunhua
and Peng, Yifan},
title={Closing the gap between open source and commercial large language models for medical evidence summarization},
journal={npj Digital Medicine},
year={2024},
month={Sep},
day={09},
volume={7},
number={1},
pages={239},
issn={2398-6352},
note={doi: 10.1038/s41746-024-01239-w},
url={https://doi.org/10.1038/s41746-024-01239-w}
}

@misc{Tang2023,
    title={Does Synthetic Data Generation of LLMs Help Clinical Text Mining?}, 
    author={Ruixiang Tang and Xiaotian Han and Xiaoqian Jiang and Xia Hu},
    year={2023},
    eprint={2303.04360},
    archivePrefix={arXiv},
    primaryClass={cs.CL},
    url={https://arxiv.org/abs/2303.04360},
    note={doi: 10.48550/arXiv.2303.04360}
}

@article{Eigenschink2023,
  author={Eigenschink, Peter and Reutterer, Thomas and Vamosi, Stefan and Vamosi, Ralf and Sun, Chang and Kalcher, Klaudius},
  journal={IEEE Access}, 
  title={Deep Generative Models for Synthetic Data: A Survey}, 
  year={2023},
  volume={11},
  number={},
  pages={47304-47320},
  note={doi: 10.1109/ACCESS.2023.3275134}
}

@article{Peng2023,
author={Peng, Cheng
and Yang, Xi
and Chen, Aokun
and Smith, Kaleb E.
and PourNejatian, Nima
and Costa, Anthony B.
and Martin, Cheryl
and Flores, Mona G.
and Zhang, Ying
and Magoc, Tanja
and Lipori, Gloria
and Mitchell, Duane A.
and Ospina, Naykky S.
and Ahmed, Mustafa M.
and Hogan, William R.
and Shenkman, Elizabeth A.
and Guo, Yi
and Bian, Jiang
and Wu, Yonghui},
title={A study of generative large language model for medical research and healthcare},
journal={npj Digital Medicine},
year={2023},
month={Nov},
day={16},
volume={6},
number={1},
pages={210},
issn={2398-6352},
note={doi: 10.1038/s41746-023-00958-w},
url={https://doi.org/10.1038/s41746-023-00958-w}
}

@article{Falis2024,
    author = {Falis, Matúš and Gema, Aryo Pradipta and Dong, Hang and Daines, Luke and Basetti, Siddharth and Holder, Michael and Penfold, Rose S and Birch, Alexandra and Alex, Beatrice},
    title = "{Can GPT-3.5 generate and code discharge summaries?}",
    journal = {Journal of the American Medical Informatics Association},
    volume = {31},
    number = {10},
    pages = {2284-2293},
    year = {2024},
    month = {09},
    issn = {1527-974X},
    note = {doi: 10.1093/jamia/ocae132},
    url = {https://doi.org/10.1093/jamia/ocae132},
    eprint = {https://academic.oup.com/jamia/article-pdf/31/10/2284/59206416/ocae132.pdf},
}

@misc{Melamud2019,
    title={Towards Automatic Generation of Shareable Synthetic Clinical Notes Using Neural Language Models}, 
    author={Oren Melamud and Chaitanya Shivade},
    year={2019},
    eprint={1905.07002},
    archivePrefix={arXiv},
    primaryClass={cs.CL},
    url={https://arxiv.org/abs/1905.07002},
    note={doi: 10.48550/arXiv.1905.07002}
}

@inproceedings{Xu2024,
    title = "Knowledge-Infused Prompting: Assessing and Advancing Clinical Text Data Generation with Large Language Models",
    author = "Xu, Ran  and
      Cui, Hejie  and
      Yu, Yue  and
      Kan, Xuan  and
      Shi, Wenqi  and
      Zhuang, Yuchen  and
      Wang, May Dongmei  and
      Jin, Wei  and
      Ho, Joyce  and
      Yang, Carl",
    editor = "Ku, Lun-Wei  and
      Martins, Andre  and
      Srikumar, Vivek",
    booktitle = "Findings of the Association for Computational Linguistics ACL 2024",
    month = aug,
    year = "2024",
    address = "Bangkok, Thailand and virtual meeting",
    publisher = "Association for Computational Linguistics",
    url = "https://aclanthology.org/2024.findings-acl.916",
    note = "doi: 10.18653/v1/2024.findings-acl.916",
    pages = "15496--15523",
}

@article{Frei2023,
    title = {Annotated dataset creation through large language models for non-english medical NLP},
    journal = {Journal of Biomedical Informatics},
    volume = {145},
    pages = {104478},
    year = {2023},
    issn = {1532-0464},
    note = {doi: 10.1016/j.jbi.2023.104478},
    url = {https://www.sciencedirect.com/science/article/pii/S1532046423001995},
    author = {Johann Frei and Frank Kramer},
}

@inproceedings{Amin-Nejad2020,
    title = "Exploring Transformer Text Generation for Medical Dataset Augmentation",
    author = "Amin-Nejad, Ali  and
      Ive, Julia  and
      Velupillai, Sumithra",
    editor = "Calzolari, Nicoletta  and
      B{\'e}chet, Fr{\'e}d{\'e}ric  and
      Blache, Philippe  and
      Choukri, Khalid  and
      Cieri, Christopher  and
      Declerck, Thierry  and
      Goggi, Sara  and
      Isahara, Hitoshi  and
      Maegaard, Bente  and
      Mariani, Joseph  and
      Mazo, H{\'e}l{\`e}ne  and
      Moreno, Asuncion  and
      Odijk, Jan  and
      Piperidis, Stelios",
    booktitle = "Proceedings of the Twelfth Language Resources and Evaluation Conference",
    month = may,
    year = "2020",
    address = "Marseille, France",
    publisher = "European Language Resources Association",
    url = "https://aclanthology.org/2020.lrec-1.578",
    pages = "4699--4708",
    language = "English",
    ISBN = "979-10-95546-34-4",
}

@misc{Lorge2024,
    title={Detecting the Clinical Features of Difficult-to-Treat Depression using Synthetic Data from Large Language Models}, 
    author={Isabelle Lorge and Dan W. Joyce and Niall Taylor and Alejo Nevado-Holgado and Andrea Cipriani and Andrey Kormilitzin},
    year={2024},
    eprint={2402.07645},
    archivePrefix={arXiv},
    primaryClass={cs.CL},
    url={https://arxiv.org/abs/2402.07645},
    note={doi: 10.48550/arXiv.2402.07645}
}

@article{Cowie2017,
    author={Cowie, Martin R.
    and Blomster, Juuso I.
    and Curtis, Lesley H.
    and Duclaux, Sylvie
    and Ford, Ian
    and Fritz, Fleur
    and Goldman, Samantha
    and Janmohamed, Salim
    and Kreuzer, J{\"o}rg
    and Leenay, Mark
    and Michel, Alexander
    and Ong, Seleen
    and Pell, Jill P.
    and Southworth, Mary Ross
    and Stough, Wendy Gattis
    and Thoenes, Martin
    and Zannad, Faiez
    and Zalewski, Andrew},
    title={Electronic health records to facilitate clinical research},
    journal={Clinical Research in Cardiology},
    year={2017},
    month={Jan},
    day={01},
    volume={106},
    number={1},
    pages={1-9},
    issn={1861-0692},
    note={doi: 10.1007/s00392-016-1025-6},
    url={https://doi.org/10.1007/s00392-016-1025-6}
}

@article{Gunter2005,
    title     = "The emergence of national electronic health record architectures
               in the United States and Australia: models, costs, and questions",
    author    = "Gunter, Tracy D and Terry, Nicolas P",
    journal   = "J. Med. Internet Res.",
    publisher = "JMIR Publications Inc.",
    volume    =  7,
    number    =  1,
    pages     = "e3",
    month     =  mar,
    year      =  2005,
    language  = "en",
    note = "\url{https://pmc.ncbi.nlm.nih.gov/articles/PMC1550638/}"
}

@misc{Dubey2024,
    title={The Llama 3 Herd of Models}, 
    author={Llama Team},
    year={2024},
    eprint={2407.21783},
    archivePrefix={arXiv},
    primaryClass={cs.AI},
    url={https://arxiv.org/abs/2407.21783},
    note={doi: 10.48550/arXiv.2407.21783}
}

@article{Nunes2024,
    title = {MediAlbertina: An European Portuguese medical language model},
    journal = {Computers in Biology and Medicine},
    volume = {182},
    pages = {109233},
    year = {2024},
    issn = {0010-4825},
    note = {doi: 10.1016/j.compbiomed.2024.109233},
    url = {https://www.sciencedirect.com/science/article/pii/S0010482524013180},
    author = {Miguel Nunes and João Boné and João C. Ferreira and Pedro Chaves and Luis B. Elvas}
}

@article{Budu2024,
    title = {Evaluation of synthetic electronic health records: A systematic review and experimental assessment},
    journal = {Neurocomputing},
    volume = {603},
    pages = {128253},
    year = {2024},
    issn = {0925-2312},
    note = {doi: 10.1016/j.neucom.2024.128253},
    url = {https://www.sciencedirect.com/science/article/pii/S0925231224010245},
    author = {Emmanuella Budu and Kobra Etminani and Amira Soliman and Thorsteinn Rögnvaldsson},
}

@inproceedings{Sabia2023,
    title={Sabi{\'a}: Portuguese Large Language Models},
    author={Ramon Pires and Hugo Abonizio and Thales Sales Almeida and Rodrigo Nogueira},
    year={2023},
    booktitle={Intelligent Systems},
    pages={226--240},
    publisher={Springer Nature Switzerland},
    address={Cham},
    isbn={978-3-031-45392-2},
}

@misc{Cabrita2023,
    title={Cabrita: closing the gap for foreign languages}, 
    author={Celio Larcher and Marcos Piau and Paulo Finardi and Pedro Gengo and Piero Esposito and Vinicius Caridá},
    year={2023},
    eprint={2308.11878},
    archivePrefix={arXiv},
    primaryClass={cs.CL}
}

@inproceedings{Gloria2024,
    title = "{G}l{\'o}r{IA}: A Generative and Open Large Language Model for {P}ortuguese",
    author = "Lopes, Ricardo  and
    Magalhaes, Joao  and
    Semedo, David",
    editor = "Gamallo, Pablo  and
    Claro, Daniela  and
    Teixeira, Ant{\'o}nio  and
    Real, Livy  and
    Garcia, Marcos  and
    Oliveira, Hugo Gon{\c{c}}alo  and
    Amaro, Raquel",
    booktitle = "Proceedings of the 16th International Conference on Computational Processing of Portuguese",
    month = mar,
    year = "2024",
    address = "Santiago de Compostela, Galicia/Spain",
    publisher = "Association for Computational Linguistics",
    url = "https://aclanthology.org/2024.propor-1.45",
    pages = "441--453",
}

@inproceedings{Dettmers2023,
author = {Dettmers, Tim and Pagnoni, Artidoro and Holtzman, Ari and Zettlemoyer, Luke},
title = {QLORA: efficient finetuning of quantized LLMs},
year = {2023},
publisher = {Curran Associates Inc.},
address = {Red Hook, NY, USA},
booktitle = {Proceedings of the 37th International Conference on Neural Information Processing Systems},
articleno = {441},
numpages = {28},
location = {New Orleans, LA, USA},
series = {NIPS '23}
}

@misc{Loshchilov2019,
    title={Decoupled Weight Decay Regularization}, 
    author={Ilya Loshchilov and Frank Hutter},
    year={2019},
    eprint={1711.05101},
    archivePrefix={arXiv},
    primaryClass={cs.LG},
    url={https://arxiv.org/abs/1711.05101},
    note={doi: 10.48550/arXiv.1711.05101}
}

@misc{SentenceBERT,
    title = {Sentence-BERT: Sentence Embeddings using Siamese BERT-Networks},
    howpublished = {\url{https://fq.pkwyx.com/default/https/aclanthology.org/D19-1410.pdf}},
    author={Nils Reimers and Iryna Gurevych},
    year={2019}
}

@article{Hernadez2023,
    author={Hernadez, Mikel and Epelde, Gorka and Alberdi, Ane and Cilla, Rodrigo and Rankin, Debbie},
    title={Synthetic Tabular Data Evaluation in the Health Domain Covering Resemblance, Utility, and Privacy Dimensions},
    journal={Methods Inf Med},
    year={2023},
    month={Jan},
    day={09},
    publisher={Georg Thieme Verlag KG},
    volume={62},
    number={S 01},
    pages={e19-e38},
    issn={0026-1270},
    note={doi: 10.1055/s-0042-1760247},
    url={http://www.thieme-connect.com/products/ejournals/abstract/10.1055/s-0042-1760247},
    language={EN}
}

@inproceedings{Schneider2021,
    author={Schneider, Elisa Terumi Rubel and de Souza, João Vitor Andrioli and Gumiel, Yohan Bonescki and Moro, Claudia and Paraiso, Emerson Cabrera},
    booktitle={2021 IEEE 34th International Symposium on Computer-Based Medical Systems (CBMS)}, 
    title={A GPT-2 Language Model for Biomedical Texts in Portuguese}, 
    year={2021},
    volume={},
    number={},
    pages={474-479},
    note={doi: 10.1109/CBMS52027.2021.00056}
}

@article{Li2022,
    title = {Neural Natural Language Processing for unstructured data in electronic health records: A review},
    journal = {Computer Science Review},
    volume = {46},
    pages = {100511},
    year = {2022},
    issn = {1574-0137},
    doi = {https://doi.org/10.1016/j.cosrev.2022.100511},
    url = {https://www.sciencedirect.com/science/article/pii/S1574013722000454},
    author = {Irene Li and Jessica Pan and Jeremy Goldwasser and Neha Verma and Wai Pan Wong and Muhammed Yavuz Nuzumlalı and Benjamin Rosand and Yixin Li and Matthew Zhang and David Chang and R. Andrew Taylor and Harlan M. Krumholz and Dragomir Radev},
    note={doi: 10.1016/j.cosrev.2022.100511}
}

@Article{Breugel2024,
author={van Breugel, Boris
and Liu, Tennison
and Oglic, Dino
and van der Schaar, Mihaela},
title={Synthetic data in biomedicine via generative artificial intelligence},
journal={Nature Reviews Bioengineering},
year={2024},
month={Dec},
day={01},
volume={2},
number={12},
pages={991-1004},
issn={2731-6092},
note={doi:10.1038/s44222-024-00245-7},
url={https://doi.org/10.1038/s44222-024-00245-7}
}

@InProceedings{Kumichev2024,
author="Kumichev, Gleb
and Blinov, Pavel
and Kuzkina, Yulia
and Goncharov, Vasily
and Zubkova, Galina
and Zenovkin, Nikolai
and Goncharov, Aleksei
and Savchenko, Andrey",
editor="Bifet, Albert
and Krilavi{\v{c}}ius, Tomas
and Miliou, Ioanna
and Nowaczyk, Slawomir",
title="MedSyn: LLM-Based Synthetic Medical Text Generation Framework",
booktitle="Machine Learning and Knowledge Discovery in Databases. Applied Data Science Track",
year="2024",
publisher="Springer Nature Switzerland",
address="Cham",
pages="215--230",
isbn="978-3-031-70381-2",
note="doi: 10.1007/978-3-031-70381-2\_14"
}

@misc{dataset_inesc_repo,
  author       = {Félix, D. and Teixeira Lopes, C.},
  title        = {Dataset of synthetic clinical notes in European Portuguese generated using an open-source large language model, along with prompting and evaluation data},
  year         = {2025},
  howpublished = {INESC TEC},
  url          = {https://doi.org/10.25747/4GC6-DK48}
}

@Article{Cohen1960,
    author={Cohen, Jacob},
    title={A coefficient of agreement for nominal scales.},
    journal={Educational and Psychological Measurement},
    year={1960},
    publisher={Sage Publications},
    address={US},
    volume={20},
    pages={37-46},
    note={doi: 10.1177/001316446002000104},
    url={https://doi.org/10.1177/001316446002000104}
}

@ARTICLE{McHugh2012,
    title    = "Interrater reliability: the kappa statistic",
    author   = "McHugh, Mary L",
    journal  = "Biochem Med (Zagreb)",
    volume   =  22,
    number   =  3,
    pages    = "276--282",
    year     =  2012,
    address  = "Croatia",
    language = "en",
    note     = "doi: 10.1177/001316446002000104"
}

@article{Sim2005,
    author = {Sim, Julius and Wright, Chris C},
    title = {The Kappa Statistic in Reliability Studies: Use, Interpretation, and Sample Size Requirements},
    journal = {Physical Therapy},
    volume = {85},
    number = {3},
    pages = {257-268},
    year = {2005},
    month = {03},
    issn = {1538-6724},
    note = {doi: 10.1093/ptj/85.3.257},
    url = {https://doi.org/10.1093/ptj/85.3.257},
    eprint = {https://academic.oup.com/ptj/article-pdf/85/3/257/31670498/ptj0257.pdf},
}

@article{Viera2005,
    title={Understanding interobserver agreement: the kappa statistic},
    author={Viera, Anthony J and Garrett, Joanne M and others},
    journal={Fam med},
    volume={37},
    number={5},
    pages={360--363},
    year={2005},
    note={\url{http://www1.cs.columbia.edu/~julia/courses/CS6998/Interrater\_agreement.Kappa\_statistic.pdf}}
}

@article{Sousa2025,
   author = {Hugo Sousa and Rúben Almeida and Purificação Silvano and Inês Cantante and Ricardo Campos and Alipio Jorge},
   note = {doi: 10.1609/aaai.v39i24.34705},
   issn = {2374-3468},
   issue = {24},
   journal = {Proceedings of the AAAI Conference on Artificial Intelligence},
   month = {4},
   pages = {25192-25200},
   title = {Enhancing Portuguese Variety Identification with Cross-Domain Approaches},
   volume = {39},
   year = {2025}
}

@Article{Waseem2025,
    author={Waseem, Hafiz Muhammad
    and Islam, Saif Ul
    and Matragkas, Nikolaos
    and Epiphaniou, Gregory
    and Arvanitis, Theodoros N.
    and Maple, Carsten},
    title={Review of generative AI for synthetic data generation: a healthcare perspective},
    journal={Artificial Intelligence Review},
    year={2025},
    month={Dec},
    day={10},
    volume={59},
    number={2},
    pages={55},
    issn={1573-7462},
    note={doi: 10.1007/s10462-025-11440-2},
    url={https://doi.org/10.1007/s10462-025-11440-2}
}
